\documentclass[conference]{IEEEtran}
\IEEEoverridecommandlockouts
\usepackage{cite}
\usepackage{amsmath,amssymb,amsfonts}
\usepackage{algorithmic}
\usepackage{graphicx}
\usepackage{textcomp}
\usepackage{xcolor}

\usepackage{tikz}
\usetikzlibrary{positioning, calc, fit}
\usepackage{pgfplots}
\usetikzlibrary{pgfplots.statistics, pgfplots.colorbrewer, pgfplots.groupplots} 
\usepackage{pgfplotstable}
\pgfplotsset{compat=1.14}
\usepackage[binary-units=true]{siunitx}
\usepackage{bm}
\usepackage[inline]{enumitem}
\usepackage{todonotes}
\usepackage{algorithm, algorithmic}
\usepackage[caption=false]{subfig}
\usepackage{makecell, array}
\usepackage{stfloats}
\usepackage{multirow}
\usepackage{pgfplotstable}
\usepackage[final]{changes}
\definechangesauthor[name={Feiyang Cai}, color=blue]{FC1}
\definechangesauthor[name={Feiyang Cai}, color=red]{FC2}

\begin{document}
	
	\title{Real-time Out-of-distribution Detection in Learning-Enabled Cyber-Physical Systems
	\thanks{The material presented in this paper is based upon work supported by the National Science Foundation (NSF) under grant numbers CNS 1739328,  the Defense Advanced Research Projects Agency (DARPA) through contract number FA8750-18-C-0089, and the Air Force Office of Scientific Research (AFOSR) DDDAS through contract number FA9550-18-1-0126. The views and conclusions contained herein are those of the authors and should not be interpreted as necessarily representing the official policies or endorsements, either expressed or implied, of AFOSR, DARPA, or NSF.}
	}
	
	\author{\IEEEauthorblockN{Feiyang Cai}
		\IEEEauthorblockA{\textit{Vanderbilt University} \\
			Nashville, TN \\
			feiyang.cai@vanderbilt.edu}
		\and
		\IEEEauthorblockN{Xenofon Koutsoukos}
		\IEEEauthorblockA{\textit{Vanderbilt University} \\
			Nashville, TN \\
			xenofon.koutsoukos@vanderbilt.edu}
	}
	
	\maketitle
	\thispagestyle{plain} \pagestyle{plain}
	
	\begin{abstract}
Cyber-physical systems (CPS) greatly benefit by using machine learning components that can handle the uncertainty and variability of the real-world. Typical components such as deep neural networks, however, introduce new types of hazards that may impact system safety. The system behavior depends on data that are available only during runtime and may be different than the data used for training. Out-of-distribution data may lead to \added{a} large error and compromise safety. The paper considers the problem of efficiently detecting out-of-distribution \added{data} in CPS control systems. Detection must be robust and limit the number of false alarms while being computational efficient for real-time monitoring. The proposed approach leverages inductive conformal prediction and anomaly detection for developing a method that has \added{a} well-calibrated false alarm rate. We use variational autoencoders and deep support vector data description to learn models that \added{can be used} efficiently compute the nonconformity of new inputs relative to the training set and enable real-time detection of out-of-distribution high-dimensional inputs. We demonstrate the method using an advanced emergency braking system and a self-driving end-to-end controller implemented in an open source simulator for self-driving cars. The simulation results show very small number of false positives and detection delay while the execution time is comparable to the execution time of the original machine learning components.
\end{abstract}

\begin{IEEEkeywords}
anomaly detection, inductive conformal prediction, out-of-distribution, self-driving vehicles.
\end{IEEEkeywords}
	\section{Introduction}
\label{sec:intro}

Learning-enabled components (LECs) such as neural networks 
are used in many classes of cyber-physical systems (CPS).
Semi-autonomous and autonomous vehicles, in particular, 
are CPS examples where LECs can play a significant role for perception, planning, and control if they are complemented with methods for analyzing and ensuring safety~\cite{tuncali2018simulation,dreossi2017compositional}. 
However, there are several characteristics of LECs that can 
complicate safety analysis. 
LECs encode knowledge in a form that is not transparent.
Deep neural networks (DNNs), for example, capture features in a multitude
of activation functions that cannot be inspected to ensure that the LEC operates 
as intended. High levels of autonomy require high-capacity models that further 
obscure the system operation. Even if an LEC is trained and tested extensively, 
it is typically characterized by a nonzero error rate. More importantly,
the error rate estimated at design-time may be different than the true error 
because of out-of-distribution data.

Since training data sets are necessarily incomplete, safety assessment at design-time 
is also incomplete. Design-time verification and analysis methods must be combined with
runtime monitoring techniques that can be used for safety assurance. In real-world CPS, the uncertainty and variability of the environment may result in data that are not similar to the data used for training. Although models such as DNNs generalize well if the training and testing data are sampled from the same distribution, out-of-distribution data may lead to large errors. Further, typical DNNs do not have the capability to \replaced{appropriately}{appropriate} estimate if an input is in- or out-of-distribution. 

An LEC is trained and tested using data available at design-time 
but must be deployed in a real system and operate under possibly different conditions. 
Testing ensures that the error is satisfactory for a large number of examples, however,
during the system operation the LEC may still encounter out-of-distribution inputs. 
The proposed approach quantifies how different are the new test data from the training data
and raises an alarm to indicate that the LEC may give a prediction with large error.
Out-of-distribution detection for CPS must be robust and limit the number of false alarms while being computational efficient for real-time monitoring.
Although the paper focuses on DNNs, the approach can be used for other LECs that are designed in a similar fashion.

Detection of out-of-distribution examples in neural networks has received considerable attention especially in the context of classification tasks in computer vision~\cite{hendrycks2016baseline,hendrycks2019oe,smith2014anomaly}. 
Such detection techniques do not take into consideration the dynamical behavior of CPS, can exhibit large number of false alarms, and cannot be applied to CPS in a straightforward manner. 
Similar techniques based on single input examples are used in mobile robotics~\cite{Richter2017SafeVN,mcallister2019robustness} where the need for methods to improve robustness is identified as an important research direction. 

The proposed approach is based on conformal prediction (CP)~\cite{vovk2005algorithmic,balasubramanian2014conformal}
and conformal anomaly detection (CAD)~\cite{laxhammar2013online}. 
The main idea of these methods is to test if a new input example
conforms to the training data set by utilizing a \textit{nonconformity measure} 
which assigns a numerical score indicating how different the input example 
is from the training data set. 
The next step is to define a $p$-value as the fraction of
observations that have nonconformity scores greater than or equal to the nonconformity scores of the training examples which is then used for estimating the confidence of the prediction for the test input. In order to use the approach online, Inductive Conformal Anomaly Detection (ICAD)
is introduced in~\cite{laxhammar2015inductive} 
where the original training set is split into 
the proper training set and the calibration set and the $p$-values are computed relative to
calibration examples. 
If a $ p$-value is smaller than a predefined anomaly threshold $\epsilon $, 
the test example can be classified as an anomaly.
An important property of the approach is that \emph{the rate of detected conformal anomalies 
is well calibrated}, that is with very high probability it is less or approximately equal
to a predefined threshold $ \epsilon \in (0,1) $~\cite{laxhammar2015inductive}. The approach is used for sequential anomaly detection of time trajectories in~\cite{laxhammar2013online} and for change-point detection in~\cite{volkhonskiy2017inductive,smith2014anomaly}. Existing methods rely on nonconformity measures computed using $k$-Nearest Neighbors and Kernel Density Estimation and cannot scale
to LECs with high-dimensional inputs used in CPS.

The main contribution of the paper is real-time detection of out-of-distribution inputs.
Our approach leverages inductive conformal prediction and anomaly detection.
In order to handle high-dimensional inputs in real-time,
we propose to compute the nonconformity scores using learned models based on 
variational autoencoders (VAEs)~\cite{DBLP:journals/corr/KingmaW13} and deep support vector data description (SVDD)~\cite{ruff2018deep}. VAEs is a generative model which allows sampling multiple examples similar to the input and computing multiple $p$-values that increase the robustness of detection. SVDD is a model trained to perform anomaly detection. In our method, it is combined with a test based on a sliding window that improves the robustness of the detection.
By using ICAD, for any valid nonconformity measure, the approach ensures that the rate of detected conformal anomalies is well calibrated. 
Further, the VAE and SVDD-based methods allow the efficient computation of the nonconformity score and the real-time detection of out-of-distribution high-dimensional inputs. It should be noted that the VAE and SVDD neural networks may exhibit \added{an} error different for out-of-distribution inputs
that is different than the testing error for in-distribution inputs. However, the robustness of the detection is improved considerably by taking into account multiple input examples and comparing with the calibration nonconformity scores. 

Another contribution of the paper is the empirical evaluation using (1) an advanced emergency braking system (AEBS)
and (2) a self-driving end-to-end controller (SDEC) implemented in CARLA~\cite{Dosovitskiy17}, an open source simulator for self-driving cars.
The AEBS  uses a perception LEC to detect the nearest front obstacle on the road and estimate the distance from the host vehicle based on camera images.  The  distance  together  with the velocity  of  the  host car  are  used  as  inputs  to  a  reinforcement  learning  controller whose  objective  is  to  comfortably stop the vehicle. Out-of-distribution inputs are generated by varying a precipitation parameter provided by CARLA which introduces visual effects that may cause large error in the distance estimation resulting to a collision. The simulation results demonstrate a very small number of false positives and a detection delay less than $\SI{1}{\second}$. For the SDEC which comes with CARLA~\cite{Dosovitskiy17}, the empirical evaluation shows that the proposed method can be used to detect a class of physically realizable  attacks in end-to-end  autonomous  driving presented in~\cite{boloor2019} \added{. The attacks} \deleted{that} are realized by painted  lines  on  the  road  \added{to} cause  the  self-driving  car  to  follow  a  target  path. For both examples, the execution time of the detection method is comparable to the execution time of the original LECs which demonstrates that the method can be used in real-time.

	\section{System Model and Problem Formulation}
\begin{figure}[!ht]
	\centering
		\begin{tikzpicture}
	\node [draw, rectangle, minimum width=1.6cm, minimum height = 0.8cm,font=\scriptsize, fill=green!40](Env) at (0.8, 2.0) {Environment};
	\node [draw, dashed, rectangle, fill=gray!20, minimum width=4.4cm, minimum height = 1.8cm, font=\scriptsize, align=center, opacity=0.5] at (4.35,2.1)(e2e){};
	\node [font=\footnotesize] at (4.35, 2.70) {End-to-end control};
	\node [draw, rectangle, minimum width=1.4cm, minimum height = 0.8cm, font=\scriptsize, fill=yellow!40] at (3.2,2.0) (perception){Perception};
	\node [draw, rectangle, fill=cyan!20, minimum width=1.4cm, minimum height = 0.8cm, font=\scriptsize, align=center] at (5.5,2.0)(RL){Control};
	\node [draw, rectangle, fill=blue!20, minimum width=1.4cm, minimum height = 0.8cm, font=\scriptsize, align=center] at (7.8,2.0)(Vehicle){Physical \\ Plant};
	
	\draw [->, line width=0.3mm, >=latex](Env) -- (perception) node[pos=0.45, above, font=\scriptsize]{};
	\draw [->, line width=0.3mm, >=latex](perception)--(RL) node[midway, above, font=\scriptsize]{};
	\draw [->, line width=0.3mm, >=latex](RL)--(Vehicle) node[midway, above, font=\scriptsize]{};
	\draw [->, line width=0.3mm, >=latex](Vehicle.south)-- ++(0.0, -0.5) -| (RL.south) node[pos=0.25, above, font=\scriptsize]{};
	\draw [->,gray, line width=0.3mm, >=latex, dashed](Vehicle.east)-- ++(0.3, 0.0) -- ++(0.0, -1.2) -| (Env.south) node[pos=0.25, above, font=\scriptsize]{};
	\end{tikzpicture}
	\caption{Simplified CPS control architecture.}
	\label{fig:cps_arch}
\end{figure}
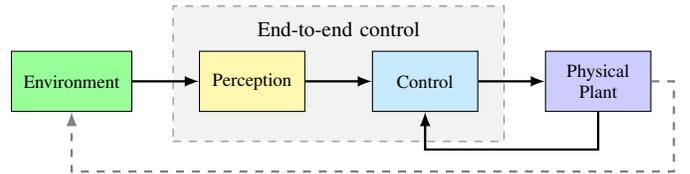

CPS use extensively LECs to perform various tasks in order to increase the level of autonomy.
A typical simplified CPS architecture with LECs (e.g., DNNs)
for perception and control is shown in Fig.~\ref{fig:cps_arch}. 
A perception component observes and interprets the environment
and provides information to a controller which, possibly using additional sensors 
(feedback from the plant), applies an action to the plant in order to achieve some task. In 
response to this action, the state of the physical plant changes and the environment 
must be observed and interpreted again in order to continue the system operation.
An end-to-end control architecture from perception to actuation can also be used. 

An LEC is designed using learning methods such as supervised and reinforcement learning. 
We assume that the LECs are successfully trained, and further, 
evaluation of training and testing errors is satisfactory. 
However, the training and testing data sets at design-time are necessarily
incomplete and may under-represent safety critical cases. 
Out-of-distribution inputs, in particular, that have not been used for training or testing
may lead to large errors and compromise safety. 

The paper considers the problem of efficiently detecting out-of-distribution 
inputs in real-time. 
The objective is to detect such input examples
in order enable decision making by switching to a different control architecture or human 
supervision. 
During the system operation, the inputs arrive one by one. After receiving each input,
the objective is to compute a valid measure of the degree to which the assumption \deleted{of} 
the input example is generated from the same probability distribution as the training data
set is falsified.

Evaluation of an online detection must be based on metrics that quantify sensitivity
and robustness. Further, out-of-distribution detection must be performed in real-time which is challenging because inputs to perception and end-to-end control
LECs are high-dimensional measurements from sensors such as cameras, LIDAR, and RADAR.
The time and memory requirements must be similar to the requirements of the LECs
used in the CPS architecture.

	\section{Background}
\label{sec:background}
\subsection{Conformal Prediction and Anomaly Detection}
\label{sec:CP}
The proposed approach is based on conformal prediction (CP)~\cite{vovk2005algorithmic,balasubramanian2014conformal}
and conformal anomaly detection (CAD)~\cite{laxhammar2013online}. 
Given a training set $ Z = \{z_1, \ldots , z_l\}$, the method aims to test if a new test example $z_{l+1}$
conforms to the training data set. 
Essential in the approach is the definition of a \textit{nonconformity measure} 
which is defined by a function $ A $ that assigns a numerical score indicating how different the test example 
is from the training data set. A large nonconformity score corresponds to a strange example relative
to the training set. There are many possible functions that can be used~\cite{balasubramanian2014conformal}. A simple example is the average distance from $z_{l+1}$ to the $k$-nearest neighbors in $Z$
($k$-nearest neighbors nonconformity measure). The next step is to define the $p$-value $p_{l+1}$
for the example $z_{l+1}$ based on the nonconformity measure. 
First, the nonconformity score $ \alpha_i$ is computed for each example $ z_i $ relative to the 
rest $\{z_1,\ldots, z_{l+1}\}\backslash z_i$ using $A$, and then, the $p$-value is defined as the fraction of
observations that have nonconformity scores greater than or equal to the nonconformity $ \alpha_{l+1}$
\begin{equation*}
p_{l+1} = \frac{|\{i=1,\ldots,l\} \,|\, \alpha_i \geq \alpha_{l+1}|}{l}.
\end{equation*}

The approach is inefficient since all
computations have to be repeated for every test example.
In order to adapt the approach for online detection, Inductive Conformal Anomaly Detection (ICAD)
is introduced in~\cite{laxhammar2015inductive} based on Inductive Conformal Prediction (ICP)~\cite{balasubramanian2014conformal}.
The original training set $(z_1, \ldots, z_l)$ is split into two sets: 
the \emph{proper training set} $(z_1, \ldots, z_m)$ and the \emph{calibration set} $(z_{m+1}, \ldots, z_l)$. 
For each example in the calibration set, the nonconformity score relative to the proper training set is precomputed by
\begin{equation*}
\alpha_i = A\Big( \{z_1,\ldots, z_{m}\}, z_i  \Big), i=m+1, \ldots, l.
\end{equation*} 
Then, given a test example $z_{l+1}$, the nonconformity score $\alpha_{l+1}$ 
is also computed relative to the \replaced{proper training}{calibration} set and the $p$-value is given by
\begin{equation}
\label{eq:p-value}
p_{l+1} = \frac{|\{i=m+1,\ldots,l \}\,|\, \alpha_i \geq \alpha_{l+1}|}{l-m}.
\end{equation}

If the $ p$-value is smaller than a predefined anomaly threshold $\epsilon \in (0,1)$, 
the test example is classified as 
a conformal anomaly. In this case, the test example can be a rare, previously unseen example 
from the same probability distribution as the training data set, an out-of-distribution example, 
or the training examples are not independent and identically distributed (IID).
The approach can be viewed as a statistical hypothesis test, where the null hypothesis
that the new example $z_{l+1}$ and the training set $(z_1, \ldots, z_l)$ are IID 
is tested at significance level $ \epsilon $. If $z_{l+1}$ and $(z_1, \ldots, z_l)$
are in fact IID, then for any choice of nonconformity measure, $\epsilon $ is an 
upper bound of the probability that $z_{l+1}$ is classified as a conformal anomaly. 
An important property of the approach is that the rate of detected conformal anomalies 
is well calibrated, that is with very high probability it is less or approximately equal
to $ \epsilon $~\cite{laxhammar2015inductive}.

Conformal prediction and anomaly detection can be applied under the 
\emph{exchangeability assumption} which is weaker than the IID assumption.
Consider a sequence $ \{z_1, z_2, \ldots , z_N\}$ generated from a probability 
distribution that is exchangeable, then for any permutation $ \pi $ of the
set $\{1,\dots,N\}$, the distribution of $\{z_{\pi(1)}, z_{\pi(2)}, \ldots , z_{\pi(N)}\}$
is the same as the distribution of the original sequence~\cite{balasubramanian2014conformal}.
Consider $ \{z_1, z_2, \ldots , z_N\}$
generated from the same exchangeable probability distribution as the proper training set. 
It is shown in~\cite{vovk2005algorithmic} (Theorem 8.2), 
that the $p$-values $ p_1, p_2, \ldots, p_N $ are independent 
and uniformly distributed in $[0,1]$. 
The sequence of $p$-values can be used to test online if the observations 
$ \{z_1, z_2, \ldots , z_N\}$  are generated from the same probability distribution 
as the proper training set since the algorithm will generate small $p$ values 
for unusual examples. 

Testing the hypothesis that $p$-values are independent 
and uniformly distributed can be performed using martingales that 
are constructed using the $p$-values~\cite{Fedorova:2012:PMT:3042573.3042693}. 
Given $ p_1, p_2, \ldots, p_N $, \cite{Fedorova:2012:PMT:3042573.3042693}
proposes the \emph{power martingale} defined for some $ \epsilon $ as
\begin{equation*}
M_{\replaced{N}{n}}^{\epsilon} = \prod_{i=1}^{\replaced{N}{n}} \epsilon p_i^{\epsilon - 1}
\end{equation*}
and the \emph{simple mixture martingale} defined as
\begin{equation}
\label{eq:martingale}
M_{\replaced{N}{n}} = \int_0^1 M_{\replaced{N}{n}}^{\epsilon} d\epsilon.
\end{equation}
Such a martingale will grow only if there are many small $p$-values in the sequence. 
If the generated $p$-values concentrate in any other part of the unit interval,  
the martingale is not expected to grow. Details and other martingales that follow \deleted{the} 
the same idea can be found in~\cite{Fedorova:2012:PMT:3042573.3042693}.

Application of the approach relies on \deleted{the} 
the nonconformal measure which must be computed efficiently and score correctly 
the strangeness of high-dimensional examples. Typical nonconformity measures such as 
the $k$-Nearest Neighbor ($k$-NN) nonconformity measure~\cite{laxhammar2015inductive} 
and the Kernel Density Estimation (KDE) nonconformity measure~\cite{smith2014anomaly}
cannot scale efficiently to high-dimensional inputs because they require 
either storing the training data set or estimating the density in a high-dimensional space. 
One of the main contributions of the paper is computing the nonconformity measures efficiently
using learned models based on variational 
autoencoders~\cite{DBLP:journals/corr/KingmaW13} and deep support vector data description~\cite{ruff2018deep}.

\subsection{Variational Autoencoders}
Variational Autoencoder (VAE) is a generative model which learns parameters of a probability distribution to represent the data~\cite{DBLP:journals/corr/KingmaW13}. 
A VAE consists of an encoder, a decoder, and a loss function.
The objective is to model the relationship between the observation 
$z$ and the low-dimensional latent variable $x$ using the loss function
\begin{equation*}
\label{eq:vaeLoss}
\mathcal{L}(\theta, \phi; z) = \mathbb{E}_{x\sim q_\phi(x|z)}[\log p_\theta(z|x)] - D_{\text{KL}}[q_\phi(x|z)||p(x)],
\end{equation*}
where $\theta$ and $\phi$ are neural network parameters. The first term is the model fit and the second is the KL divergence between the approximate posterior and the prior of $x$. A popular choice for the prior is Gaussian distribution. 
VAE-based methods can utilize the reconstruction error or reconstruction accuracy for anomaly detection~\cite{an2015variational}. 
In our approach, given an input $z$
the generative model (decoder) is used to sample IID examples from $ q_\phi(x|z)$ 
and the reconstruction accuracy is used as a nonconformity measure.

\subsection{Deep Support Vector Data Description}
VAEs and other autoencoder architectures are trained to perform a task other than anomaly
detection assuming that the reconstruction accuracy will 
be better for in-distribution examples. Deep Support Vector Description 
(SVDD) is an architecture trained to perform anomaly detection~\cite{ruff2018deep}.
The idea is to train a DNN to  map the input data into a hypersphere of minimum
volume characterized by center $ \bm{c} $ and radius $ R $.
The input space $\mathcal{Z}$ is transformed to a compressed output space $\mathcal{X}$ while minimizing the volume of the hypersphere that encloses most of the input representations. 
Given a training data set $\{ z_1, \ldots, z_n\}$, 
\deleted{hyperparameter $\lambda>0$, where $|| \cdot ||_F$ is the Frobenius norm.
	$\phi(\cdot\,; \mathcal{W}) : \mathcal{X} \rightarrow \mathcal{F}$ denotes the neural network 
	with weights $\mathcal{W}$,
	$\bm{c}\in \mathcal{F}$ is the center of the hypersphere, and $\nu \in (0,1)$ is a parameter to }
the \textit{one-class deep SVDD}~\cite{ruff2018deep}
is based on the loss
\begin{equation*}
\label{eq:one-class}
\min_{\mathcal{W}}\quad \frac{1}{n}\sum_{i=1}^{n}||\phi(z_i;\mathcal{W})- \bm{c}||^2+\frac{\lambda}{2}\sum_{\ell=1}^{L}|| \bm{W}^\ell ||^2_F,
\end{equation*}
where $\phi(\cdot\,; \mathcal{W}) : \mathcal{Z} \rightarrow \mathcal{X}$ denotes the neural network with weights $\mathcal{W}$, $\bm{c}\in \mathcal{X}$ is the center of the hypersphere, and
the last term is a weight regularizer with 
hyperparameter $\lambda>0$, where $|| \cdot ||_F$ is the Frobenius norm.
One-class deep SVDD learns to map the data as close to center $\bm{c}$ as possible by 
penalizing the distance from representations to the center. 
The deep SVDD neural network must not have bias terms or bounded activation 
functions and the center $\bm{c}$ can be selected as the mean of the 
representations from the initial inference on some training data to avoid
trivial solutions that map the input space to a single point~\cite{ruff2018deep}. 
Given a new test example $z$, the distance of the representation 
$\phi(z; \mathcal{W}^*)$ 
to the center $\bm{c}$ of the hypersphere reflects how different the test example is 
from the training data set and can be used as a nonconformity measure.

	\section{Out-of-distribution Detection}
\label{sec:method}
 

\subsection{Detection Algorithm}
The algorithm is based on the ICP and ICAD, and therefore, the training data set is split
into a proper training set $(z_1, \ldots, z_m)$ and a calibration set  
$(z_{m+1},  \ldots, z_l)$. Practically, the data set used for training the LEC 
can be used as the proper training set, and after the LEC training phase,
additional data can be collected to form the calibration set.
For each example in the calibration set, a function $ A $ is used to compute the 
nonconformity measure. The nonconformity scores of the calibration 
examples are sorted in order to be used at runtime. 
Algorithm~\ref{alg:offline} shows the steps that are performed offline.

\begin{algorithm}[!ht]
	\caption{Offline algorithm for computing the calibration nonconformity scores}
	\label{alg:offline}
	\begin{algorithmic}[1]
		\renewcommand{\algorithmicrequire}{\textbf{Input:}}
		\renewcommand{\algorithmicensure}{\textbf{Output:}}
		\REQUIRE Training set $(z_1,\ldots, z_l)$, number of calibration examples $l-m$
		\ENSURE  Sorted list of calibration nonconformity scores ($\alpha_{m+1}, \ldots, \alpha_{l}$)
		\STATE Split the training set into the proper training set $(z_1,\ldots, z_{m})$ and calibration set $(z_{m+1},\ldots, z_{l})$
		\FOR {$j = m+1$ to $l$}
		\STATE $ \alpha^\prime_j = A\Big( \{z_1,\ldots, z_{m}\}, z_j \Big) $
		\ENDFOR
		\STATE ($\alpha_{m+1}, \ldots, \alpha_{l}$) = sort($\alpha^\prime_{m+1}, \ldots, \alpha^\prime_{l}$)
	\end{algorithmic} 
\end{algorithm}

Let us consider an LEC defining a mapping from input $z$ to output $y$. 
The set of input examples used for training is denoted by $(z_1,  \ldots, z_l)$.
During the system operation, a sequence of inputs denoted by
($z^\prime_1, \ldots, z^\prime_{N}$) is processed one-by-one.
The task of the algorithm is to quantify how different the input sequence 
is from the training data set. If the difference is large, the algorithm raises 
an alarm indicating that the LEC may generate an output $ y $ with large error
compared to the testing error obtained at design-time.

At runtime, given a new input $z^\prime_k$, the noncomformity score $\alpha^\prime_k$ 
can be computed using the nonconformity function $ A $ relative to the proper training set
\begin{equation*}
\alpha^\prime_k = A\Big( \{z_1,\ldots, z_{m}\}, z^\prime_k \Big).
\end{equation*} 
The computation requires evaluating the strangeness of $ z^\prime_k $ relative to 
$ \{z_1,\ldots, z_{m}\} $.
The choice of the nonconformity function $A$ must ensure computing informative
nonconformity scores in real-time. Using, for example, 
$k$-NN requires storing the training data set which is infeasible 
for real-world CPS. Instead, we propose to learn an appropriate neural network architecture
which is trained offline using the proper training set and encodes the required information
in its parameters. This neural network \emph{monitors} the inputs to the perception or 
end-to-end control LEC and is used to compute in real-time the nonconformity measure. 

Given an input $ z^\prime_k $, the $p$-value $p_k$ is computed as the fraction of 
calibration examples that have 
nonconformity scores greater than or equal to $\alpha^\prime_k$ using Eq. (\ref{eq:p-value}). 
It should be noted that the computation of the $p$-value can be performed efficiently 
online since it requires storing only the calibration nonconformity scores.
If $ p_k < \epsilon $ the example $ z^\prime_k $ 
is classified as an anomaly. Using a single $p$-value for detecting
out-of-distribution examples can lead to an oversensitive detector with a large
number of false alarms that inhibit the operation of the CPS. Our objective 
is to compute a sequence of $p$-values and use the martingale-based method presented 
in Section~\ref{sec:background} to test if the $p$-values are independent and 
uniformly distributed indicating that the input is generated from the same 
probability distribution as the proper training data set or there are many small
$p$-values indicating an out-of-distribution input. In the latter case, the martingale
will grow and can be used as an input to a detector that raises alarms.

In the following, we describe how VAE and SVDD are used to compute the nonconformity 
measure and detect out-of distribution examples.
Although the idea is similar, the two architectures lead to different algorithms for 
computing the sequence of $p$-values and realizing the detector.

\subsection{VAE-based Out-of-distribution Detection}

\subsubsection{Nonconformity measure and $p$-values}
Given an input example $z_t$ at time $t$, the encoder portion of a VAE is used
to approximate the posterior distribution of the latent space and 
sample multiple points $x_k$ from the posterior that are used as input to the decoder portion in order to
and generate new examples $z^\prime_1, \dots, z^\prime_N$. 
Typically, the posterior of the latent space is approximated by a Gaussian distribution. 
Sampling from the posterior generates encodings $x_k$ 
so that the decoder is exposed to a range \replaced{of} {a} variations of the input example and 
outputs $z^\prime_1, \dots, z^\prime_N$ which satisfy the exchangeability assumption.

An in-distribution input $ z_t $ should be reconstructed with a relatively small reconstruction error. Conversely, an out-of-distribution input will \replaced{likely have}{have likely} a larger
error. The reconstruction error is a good evaluation of the strangeness of the input
relative to the training set and it is used as the nonconformity measure.
We use the squared error between the input example $ z_t $ and each generated output 
example $ z^\prime_k $ as the nonconformity \replaced{measure}{measured} defined as
\begin{equation}
\label{eq:vae-ncm}
    \alpha^\prime_k = A_{\text{VAE}}(z_t,z^\prime_k) = || z_t - z^\prime_k ||^2.
\end{equation}

The $p$-value $p_k$ for the input $ z^\prime_k $  is computed as the fraction of 
calibration examples that have 
nonconformity scores greater than or equal to $\alpha^\prime_k$ using Eq. (\ref{eq:p-value}).
Since the examples $z^\prime_1, \dots, z^\prime_N$ satisfy the
exchangeability assumption, the $p_k$ values are independent and uniformly 
distributed in $[0,1]$ (see discussion \deleted{and references} in Section~\ref{sec:background})
and the martingale method can be used to test if $z^\prime_1, \dots, z^\prime_N$, and
therefore $ z_t $, are generated from the probability distribution of the training data.

\subsubsection{Martingale test}
At runtime, for every new input example $ z_t $ 
received by the perception or end-to-end control LEC at time $t$ we compute the martingale
\begin{equation*}
M_{t} = \int_0^1 M_{t}^{\epsilon} d\epsilon \\
	  = \int_0^1 \prod_{k=1}^N \epsilon p_k^{\epsilon - 1} d\epsilon.
\end{equation*}
$ M_t$ will have a large value if there are many small $p$-values in the sequence
$ p_k $ which will indicate an out-of-distribution input.

\subsubsection{Stateful detector}
In order to robustly detect when $M_t$ becomes consistently large, we use the Cumulative
sum (CUSUM) procedure~\cite{basseville1993detection}. CUSUM is a nonparametric stateful test 
and can be used to generate alarms for out-of-distribution inputs by keeping track of the historical information of the martingale values. 

The detector is defined as $S_1=0$ and $S_{t} = \max(0, S_{t-1} + M_{t-1} - \delta)$,  
where $\delta$ prevents $S_t$ from increasing consistently when the inputs are 
in the same distribution as the training data. An alarm is raised whenever $S_t$ is 
greater than a threshold $S_t > \tau $ which can be optimized using empirical
data~\cite{basseville1993detection}. Typically, after an alarm the test is 
reset with $ S_{t+1} = 0$.

Algorithm~\ref{alg:vae} describes the VAE-based real-time out-of-distribution detection.
The nonconformity measure can be computed very efficiently by executing the learned 
VAE neural network and generating $ N $ new examples. The complexity is comparable 
to the complexity of the perception or end-to-end LEC that is executed in real-time.

\begin{algorithm}[!ht]
	\caption{VAE-based out-distribution detection}
	\label{alg:vae}
	\begin{algorithmic}[1]
		\renewcommand{\algorithmicrequire}{\textbf{Input:}}
		\renewcommand{\algorithmicensure}{\textbf{Output:}}
		\REQUIRE Input example $z_t$,
		calibration nonconformity scores ($\alpha_{m+1}, \ldots, \alpha_{l}$), 
		number of examples to be sampled $N$, 
		stateful detector threshold $\tau$ and parameter $\delta$
		\ENSURE  Output boolean variable $Anom_{t}$
		\FOR {$t = 1, 2, \ldots$}
		\FOR {$k = 1$ to $N$}
		\STATE Sample $ z^\prime_k $ using the trained VAE
		\STATE $\alpha^\prime_{k} = A_{\text{VAE}}(z_t,z^\prime_k)$
		\STATE $p_k, = \frac{|\{i=m+1,\ldots,l\} \,|\, \alpha_i \geq \alpha^\prime_k|}{l - m}$ 
		\ENDFOR
		\STATE $M_t =  \int_0^1 \prod_{k=1}^N \epsilon p_k^{\epsilon - 1} d\epsilon$
		\IF {$t=1$} 
		\STATE $S_t = 0$
		\ELSE
		\STATE $S_t = \max(0, S_{t-1} + M_{t-1} -\delta)$
		\ENDIF
		\STATE $Anom_{t} \leftarrow S_t > \tau $
		\ENDFOR
	\end{algorithmic} 
\end{algorithm}

\subsection{SVDD-based out-of-distribution Detection}

\subsubsection{Nonconformity measure and $p$-values}
The SVDD-based method also uses a 
learned model to calculate the nonconformity score. 
The proper training set is used to train the deep SVDD model. The center of the hypersphere
$\bm{c}$ is fixed as the mean of the representations from the initial pass on the 
proper training data.
After training, the neural network function $\phi(z_t, \mathcal{W^*})$ maps 
an input example $z_t$ to a representation close to the center $\bm{c}$.
In-distribution inputs are likely concentrated in a relatively small area in the
output space while the out-of-distribution inputs will be faraway from the center. 
The distance of the representation to the center $\bm{c}$ of the hypersphere 
can be used to evaluate the strangeness of the test example relative to the proper 
training set and is defined as the nonconformity measure
\begin{equation*}
    \alpha^\prime_t = A_{\text{SVDD}}(z_t) = || \phi(z_t;\mathcal{W}^*) - \bm{c}||^2 .
\end{equation*}
The $p$-value is computed as the fraction of calibration examples that have 
nonconformity scores greater than or equal to $\alpha^\prime_t$ (Eq.(\ref{eq:p-value})).
However, in contrast to the VAE, SVDD is not a generative model and cannot be used 
to generate multiple IID examples similar to $ z_t $. 

\subsubsection{Martingale test}
In order to improve the robustness of out-of-distribution detection, it is desirable
to use a sequence of inputs. In CPS, the inputs arrive at the perception or end-to-end LEC one-by-one and they are time-correlated, and therefore not independent. For a sequence of 
inputs $z_t: t = 1,2,\ldots $, the martingale in Eq. (\ref{eq:martingale}) will 
increase continuously even for in-distribution examples. In order to adapt the test, 
we use a sliding window $ [t-N+1,t] $, and given an input sequence 
$(z_{t-N+1}, \ldots, z_t)$, we compute the sequence of $p$-values  $(p_{t-N+1}, \ldots, p_t)$.
Although the $p$-values are not guaranteed to be independent and uniformly distributed, out-of-distribution inputs will still result in small $p$-values and the martingale test 
is used to identify sequences with many small values. 
In this case, the martingale is given by
\begin{equation*}
M_t = \int_0^1 M_t^{\epsilon} d\epsilon \\
= \int_0^1 \prod_{i=t-N+1}^t \epsilon p_{i}^{\epsilon - 1} d\epsilon.
\end{equation*}

In order to apply this method to CPS, the rate that we receive observations 
from the environment must be much faster than the dynamic evolution of the system and the main
factor that differentiates consecutive observations are random disturbances and noise.
For a short window, it can be assumed that the input sequence 
$(z_{t-N+1}, \ldots, z_t)$ satisfies the exchangeability assumption
and the martingale test can be used to detect multiple small $p$-values in a short
time interval. It should be noted that the martingale $ M_t $ does not depend on the
order of the input examples $(z_{t-N+1}, \ldots, z_t)$ . Also, $ M_t $ must be initialized
for the first steps using, for example, random independent and uniformly distributed  $p$-values.
 
\subsubsection{Stateless detector}
Since we already use a sliding window to compute $ M_t$, we employ a stateless
detector based on the value $M_t$ and a predefined thershold $\tau $ expressed as 
$ M_t > \tau $.

Algorithm~\ref{alg:svdd} describes the SVDD-based real-time out-of-distribution detection.
Compared with the VAE, the SVDD based method is more efficient since it does not require
generating multiple examples at each step. The martingale $M_t$ can be computed recursively
by incorporating the $p$-value for the new input and omitting the last \replaced{one}{$p$-value} in the
sliding window.

\begin{algorithm}[!ht]
	\caption{SVDD-based real-time out-distribution detection}
	\label{alg:svdd}
	\begin{algorithmic}[1]
		\renewcommand{\algorithmicrequire}{\textbf{Input:}}
		\renewcommand{\algorithmicensure}{\textbf{Output:}}
		\REQUIRE Input example $z_t$,
		calibration nonconformity scores ($\alpha_{m+1}, \ldots, \alpha_{l}$),
		sliding window size $N$, stateless detector threshold $\tau$
		\ENSURE  Output boolean variable $Anom_{t}$
		\FOR {$t = 1, 2, \ldots$}
		\STATE $\alpha^\prime_t = A_{\text{SVDD}}(z_t)$
		\STATE $p_t = \frac{|\{i=m+1,\ldots,l \}\,|\, \alpha_i \geq \alpha^\prime_{t}|}{l - m}$ 
		\STATE $M_t =  \int_0^1 \prod_{i=t-N+1}^t \epsilon p_{i}^{\epsilon - 1} d\epsilon$
		\STATE $Anom_{t} \leftarrow M_t > \tau $
		\ENDFOR
	\end{algorithmic} 
\end{algorithm}

	\section{Evaluation}
We evaluate the proposed approach using (1) an advanced emergency braking system (AEBS)
and (2) a self-driving end-to-end controller (SDEC).
The AEBS and SDEC are implemented using CARLA~\cite{Dosovitskiy17}. 
We use CARLA 0.9.5 on a $16$-core i7 desktop with $\SI{32}{\giga\byte}$ RAM memory and a single RTX 2080 GPU with $\SI{8}{\giga\byte}$ video memory. 

\subsection{Advanced Emergency Braking System}

\subsubsection{Experimental Setup}
\begin{figure}[!ht]
 \centering
 \input{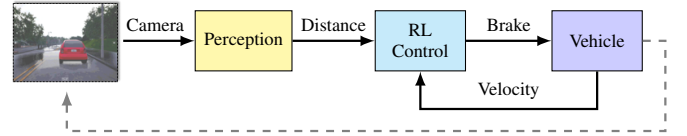}
 \caption{Advanced emergency braking system architecture.}
 \label{fig:aebs_architecture}
\end{figure}

 The architecture of the AEBS is shown in Fig.~\ref{fig:aebs_architecture}. A perception LEC
 is used to detect the nearest front obstacle on the road and estimate the distance. The distance together with the velocity of the host car are used as inputs to a reinforcement learning controller whose objective is to generate the appropriate braking force in order to safely and comfortably stop the vehicle.

\begin{figure}[!ht]
	\centering
	\input{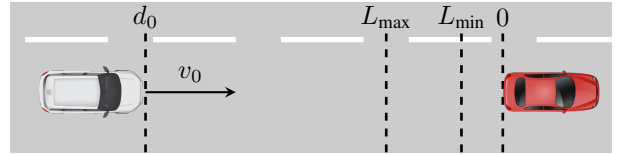}
	\caption{Illustration of advanced emergency braking system.}
	\label{fig:aebs_senario}
\end{figure}

The desirable behavior is illustrated in Fig.~\ref{fig:aebs_senario}. The AEBS detects a stopped lead car and applies the brake to decelerate and avoid the potential collision. The initial velocity of the host vehicle is $v_0$ and the initial distance between the host car and the obstacle is $d_0$. The goal of the controller is to stop the car between $L_{\min}$ and $L_{\max}$.
The sampling period used in the simulation is $\Delta t = \SI{1/20}{\second}$. In order to
simulate realistic scenarios, we introduce uncertainty into the system. The initial velocity $v_0$ is uniformly sampled between $\SI[per-mode=symbol]{90}{\kilo\meter\per\hour}$ and $\SI[per-mode=symbol]{100}{\kilo\meter\per\hour}$, and the initial distance $d_0$ is 
\deleted{based on the camera range in CARLA and is} approximately $\SI{100}{\meter}$. 
CARLA allows controlling the precipitation in the simulation using a parameter which takes values in $ [0,100]$. For training the perception LEC, and also the VAE and SVDD used for out-of-distribution detection, the precipitation parameter is randomly sampled from the interval $[0,20]$. The uncertainty introduced affects the error of the perception LEC. It should be noted that this is just a visual effect and it does not affect the physical behavior of the car.

The perception LEC is implemented using a convolutional neural network (CNN) which is trained using 
supervised learning with a training data set consisting of $8160$ images obtained by varying the 
simulation parameters described above. 
The perception LEC has three layers of $ 24/36/48 \times (5 \times 5)$ filters with ReLU activations and $2\times2$ strides, two layers of $64/64 \times (3\times3)$ filters with ReLU activations and $1 \times 1$ strides, three fully connected layers of $100/50/10$ units with ReLU activations and an output layer of size $1$ with Sigmoid activation. After $100$-epoch training, the mean absolute errors for training and testing are $\SI{0.54}{\meter}$ and $\SI{0.56}{\meter}$ respectively and are used to select $L_{\min}$ and ensure safety.
The reinforcement learning controller is trained using the DDPG algorithm~\cite{DBLP:journals/corr/LillicrapHPHETS15} with $1000$ episodes and reward function which aims to stop the vehicle between $L_{\min}=\SI{1}{\meter}$ and $L_{\max}=\SI{3}{\meter}$. A simulation run is shown in Fig.~\ref{fig:InDist}. Initially, the distance between the host and the lead car is $\SI{97.56}{\meter}$, and the velocity of the host car is $\SI[per-mode=symbol]{96.84}{\kilo\meter\per\hour}$ ($=\SI[per-mode=symbol]{26.90}{\meter\per\second}$). 
After $140$ steps or $\SI{7.00}{\second}$, the host vehicle stops at 
$\SI{1.98}{\meter}$ from the lead car. 

\subsubsection{VAE and SVDD training}
The data set with the $8160$ images used for training the perception LEC is used as the proper training data set. In addition, using simulations with the same random parameters, we collect $2040$ images for the calibration set.
We use a VAE with four layers of $ 32/64/128/256 \times (5\times5)$ filters with exponential linear unit (ELU) activations and $2\times2$ max-pooling, one fully connected layer of size $1568$ with ELU activation, $1024$ latent space, and a symmetric deconvolutional decoder. A simple two-phase learning schedule is employed with initial searching learning rate $\eta=10^{-4}$ for $250$ epochs, and subsequently fine-tuning $\eta=10^{-5}$ for $100$ epochs.

The Deep SVDD is similar with four convolutional layers of  $ 32/64/128/256 \times (5\times5)$ filters with ELU activations and $2\times2$ max-pooling, followed by one fully connected layer of $1568$ units. As suggested in~\cite{ruff2018deep}, we first train a deep convolutional autoencoder (DCAE) to initialize the deep SVDD. After $250 \ (\eta=10^{-4}) + 100\ (\eta=10^{-5})$ epochs of DCAE training, we copy the weights to the SVDD and set the hypersphere center $\bm{c}$ to the mean of the reduced space of the initial forward inference. The one-class deep SVDD objective is used as the loss and the neural network is trained for additional $150 \ (\eta=10^{-4}) + 100\ (\eta=10^{-5})$ epochs.



  
\subsubsection{Results}
To characterize the performance of the out-of-distribution detection, we use multiple simulation episodes that include in- and out-of-distribution examples. Each episode starts with a random initial velocity $ v_0 $ of the host car. The AEBS is activated upon detection of the lead car by the camera as implemented in CARLA.  We vary the precipitation parameter $ r$ as
\begin{equation*}
	r = \left\{ \begin{array}{lcl}
	r_0 & \mbox{for} & t<t_0 \\ 
	r_0+\beta (t-t_0) & \mbox{for} & t_0\leq t \leq t_1 \\ 
	r_0+\beta (t_1-t_0) & \mbox{for} & t>t_1 \end{array}\right.
\end{equation*}
where $r_0$ is the initial precipitation uniformly sampled from $[0,10]$; 
$t_0\in\{10,11, \ldots, 30\}$ is selected randomly as the time step the precipitation starts to increase; $t_1\in\{90,91, \ldots, 110\}$ is selected randomly as the time step the precipitation stops increasing; and $\beta\in[0.1,0.5]$ is a randomly selected slope. In some episodes $ r$ is always below $20$ (in-distribution) while in other episodes $ r $ exceeds $ 20 $ and it is assumed that the perception LEC receives out-of-distribution inputs. We simulate $200$ episodes and $108$ of them are in-distribution while $92$ of them contain out-of-distribution inputs.

\begin{figure}[!ht]
	\centering
	\input{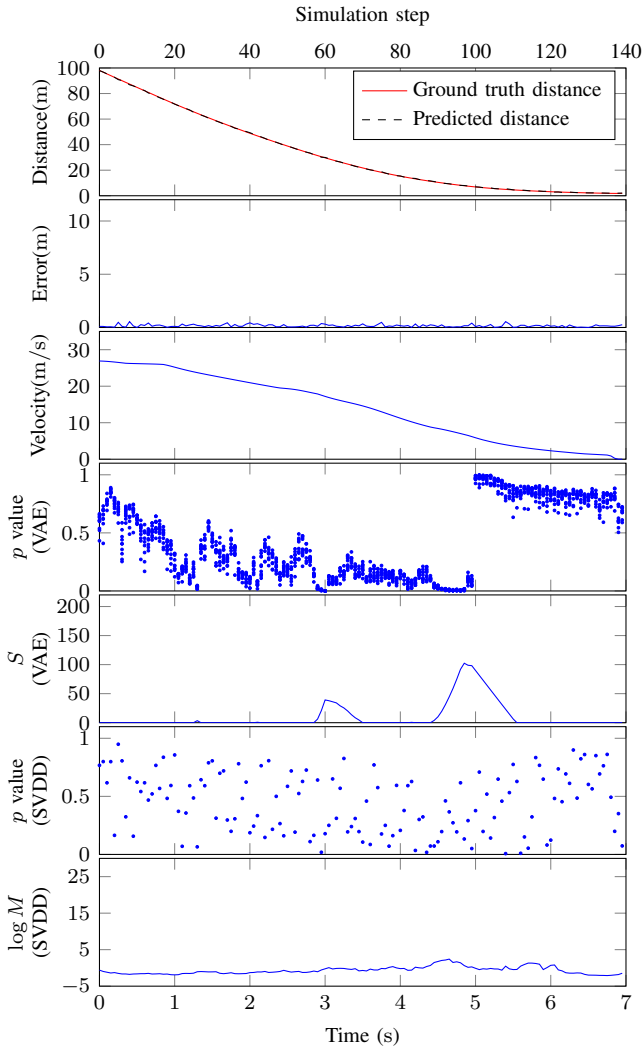}
	\caption{Episode with in-distribution inputs.}
	\label{fig:InDist}
\end{figure}

We illustrate the approach using two episodes \replaced{and}{an} we plot the 
ground truth and the predicted distance to the lead car, the velocity of the host car, the $p$-value and stateful detector $S$-value computed using the logarithm of $ M_t $ and $\delta=6$, the $p$-value of the SVDD-based method, and the logarithm of the SVDD-based martingale. Since $ M_t $ takes very large values, $\log M_t $ is used.
We use $N=10$ for the number of examples generated by the VAE and the size of sliding window in the SVDD-based approach. 
Fig.~\ref{fig:InDist} shows simulation results for the in-distribution case. 
The $p$-values are randomly distributed between $0$ to $1$, and the martingale for both approaches is small. The VAE-based  method is more sensitive than the SVDD as indicated by the larger value around $\SI{5}{\second}$. The $p$-values may become small if the camera observations are different than the training data. In this case, there is a speed limit traffic sign that is not accurately reconstructed. After the car passes the traffic sign, the $p$-values increase and the martingale goes back to very small value. The effect in the SVDD method is smaller, since we use a sliding window. Even if the training data set is augmented, there will be always novelties that can lead to variation but a robust detection method can limit the number of false alarms.

\begin{figure}[!ht]
	\centering
	\input{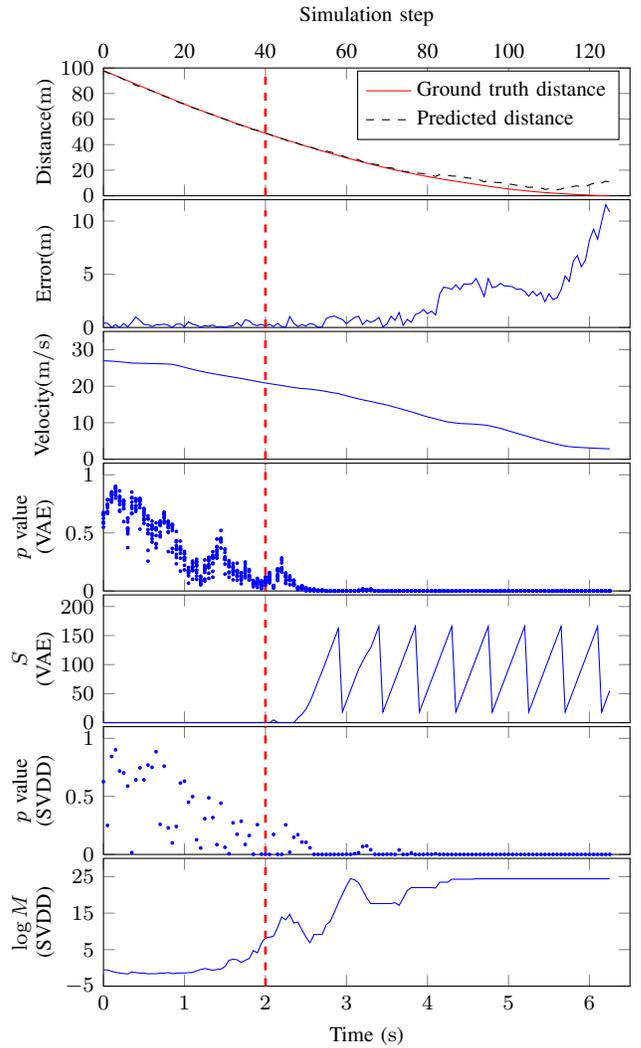}
	\caption{Episode with out-of-distribution inputs.}
	\label{fig:OutDist}
\end{figure}

An episode with out-of-distribution inputs is shown in Fig.~\ref{fig:OutDist}.  The parameter $ r $ exceeds $ 20 $ at time step $40$ $(\SI{2.0}{\second})$. 
The error of the perception LEC starts increasing and reaches almost $ \SI{11}{\meter} $. The controller is misled by the perception LEC, and does not stop the car which collides with the lead car (velocity is greater than $0$ when ground truth distance comes to $0$). 
Both VAE-based and SVDD-based martingale grow as the $p$-values become smaller. 

We evaluate the approach for the $200$ \replaced{ episodes generated by considering different values of $N$}{episodes generated by consider different values of  $ N $}. 
We run each episode and if an alarm is raised, we stop the simulation, and we check if the alarm is false\added{.} \deleted{and compute the detection delay as the number of frames from the time $ r $ exceed $ 20 $.} \added{We compute the detection delay as the number of frames from the time $ r $ exceed $ 20 $.}
We select the detector parameters $\tau$ and $\delta$ using a simple search for achieving average detection delay less than $25$ frames. Tables~\ref{tab:vae-based AEBS} and \ref{tab:svdd-based AEBS} shows the results for the VAE and SVDD-based methods respectively.
The number of false alarms is very small and the delay for detection is smaller than $ 20$ frames or $ \SI{1}{\second} $ \deleted{(the sampling rate is $\SI{20}{\hertz}$)}.

\begin{table}[!ht]
	\centering
	\caption{VAE-based detection.}
	\label{tab:vae-based AEBS}
	\begin{tabular}{c c c c}
		\hline
		\multicolumn{1}{m{0.18\columnwidth}}{\centering Parameters $(N, \delta, \tau)$} 
		 & \multicolumn{1}{m{0.20\columnwidth}}{\centering False positive}
		 & \multicolumn{1}{m{0.20\columnwidth}}{\centering False negative}    
		 & \multicolumn{1}{m{0.21\columnwidth}}{\centering Average delay (frames)}  \\
		
		\hline
		$5, 5, 42$ & $2/108$ & $0/92$ & $17.91$\\
		\hline
		$5, 5, 49$ & $0/108$ & $0/92$ & $19.84$\\
		
		\hline
		$10, 6, 156$ & $0/108$ & $0/92$ & $18.65$\\
		\hline
		$10, 10, 106$ & $0/108$ & $0/92$ & $19.30$\\
		
		\hline
		$20, 16, 250$ & $2/108$ & $0/92$ & $17.63$\\
		\hline
		$20, 18, 240$ & $0/108$ & $0/92$ & $18.46$\\
		\hline
	\end{tabular}
\end{table}

\begin{table}[!ht]
	\centering
	\caption{SVDD-based detection.}
	\label{tab:svdd-based AEBS}
	\begin{tabular}{c c c c}
		\hline
		\multicolumn{1}{m{0.18\columnwidth}}{\centering Parameters $(N, \tau)$} 
		& \multicolumn{1}{m{0.20\columnwidth}}{\centering False positive}
		& \multicolumn{1}{m{0.20\columnwidth}}{\centering False negative}    
		& \multicolumn{1}{m{0.21\columnwidth}}{\centering Average delay (frames)}  \\
		
		\hline
		$5, 8$ & $2/108$ & $0/92$ & $15.12$\\
		\hline
		$5, 9$ & $1/108$ & $0/92$ & $20.85$\\
		
		\hline
		$10, 12$ & $1/108$ & $0/92$ & $14.38$\\
		\hline
		$10, 14$ & $0/108$ & $0/92$ & $17.78$\\
		
		\hline
		$20, 16$ & $1/108$ & $0/92$ & $12.02$\\
		\hline
		$20, 17$ & $0/108$ & $0/92$ & $13.29$\\
		\hline
	\end{tabular}
\end{table}

\subsection{Self-driving End-to-end Control}

\subsubsection{Experimental Setup}
\begin{figure}[!ht]
	\centering
	\subfloat[Original image\label{fig:PhysicalAttack_a}]{
		\includegraphics[width=0.35\linewidth]{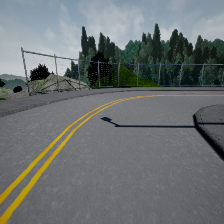}}
	\hfil
	\subfloat[Attacked image\label{fig:PhysicalAttack_b}]{
		\includegraphics[width=0.35\linewidth]{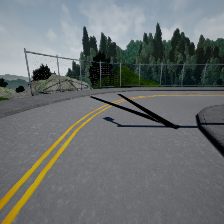}}
	\caption{Comparison of original image and image with attack~\cite{boloor2019}.}
	\label{fig:PhysicalAttack}
\end{figure}

The CARLA simulator comes with a self-driving end-to-end controller trained using imitation learning. The SDEC uses camera images as inputs and computes steering, acceleration, and brake actuation signals applied to the car. 
%
The SDEC is implemented using a CNN trained by conditional imitation learning with 14 hours of driving data recorded by human drivers~\cite{Dosovitskiy17}.
The sampling period used here is $\Delta t =  \SI{1/10}{\second}$.
For this example, our objective is to evaluate if the method can be used 
to detect a class of adversarial attacks. An approach for designing physically realizable attacks in end-to-end autonomous driving is presented in~\cite{boloor2019} and a novel class of hijacking attacks is introduced where painted lines on the road cause the self-driving car to follow a target path. Fig.~\ref{fig:PhysicalAttack_b} shows an image with the painted pattern on the road. 

In order to train the VAE and SVDD, we collect training data using episodes without attacks. We generate $633$ images in two different weather patterns (clear noon and cloudy noon) and three different scenarios (turning right, turning left, and going straight). We randomly split the training data into $506$ images for the proper training data set and $127$ images for the calibration set. We use the same VAE and SVDD architectures and hyperparameters as in the AEBS.

The evaluation focuses on the \textit{Right Corner Driving} case which is reported as more vulnerable~\cite{boloor2019}. We run $105$ simulation episodes described in~\cite{boloor2019}  with different \replaced{attacks}{attackd} such as positions and rotations of the two black lines which are chosen to cause traffic infractions.
In $69$ out of the $105$ episodes the attack is successful causing a vehicle crash. 
Our approach detects the attacks in all $105$ episodes. 
Distinguishing between attacks that cause a crash or not using only the input images is an interesting question for further research. Also, it is important to investigate how to design attacks that hijack both the end-to-end control and the out-of-distribution detection.

\begin{figure}[!ht]
	\centering
	\input{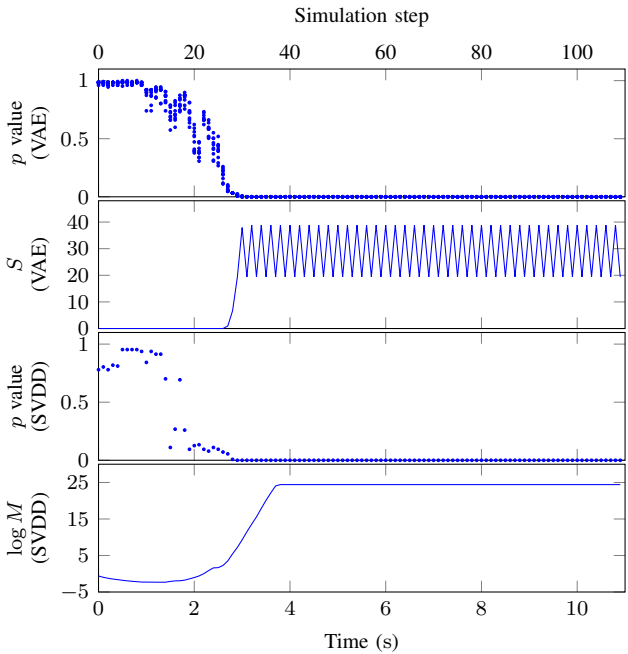}
	\caption{Episode with attacked inputs.}
	\label{fig:PhysicalOut}
\end{figure}

We plot the $p$-values and stateful detector $S$-value of the VAE-based method, the $p$-values  and the logarithm of the SVDD-based martingale in Fig.~\ref{fig:PhysicalOut} ($N=10$ and $\delta=1$). 
In this episode, there are two black lines painted on the road as shown in Fig.~\ref{fig:PhysicalAttack_b} and the vehicle is misled leading to a crash. The $p$-values are almost $0$ and the martingales grow very large indicating the input images are out-of-distribution.

\subsection{Computational Efficiency}
The VAE-based and SVDD-based methods can compute the nonconformity scores in real-time without storing training data.
Table~\ref{tab:real-time} reports the minimum $(\min)$, first quartile $(Q_1)$, second quartile or median $(Q_2)$, third quartile $(Q_3)$, and maximum $(\max)$ of (1) the execution times of the LECs in AEBS and SDEC and (2) the execution times of the VAE-based and SVDD-based detectors for different values of $N$.
Since the VAE-based method uses $N$ examples in each time step to compute the nonconformity scores, the execution time is larger than the execution time of the SVDD-based method.
The execution time of SVDD-based detection method is independent of the window size $N$ since the martingale can be computed recursively for the sliding window.
The execution times are similar to the execution times of the perception and end-to-end control LECs and much smaller than the corresponding sampling time ($\SI{50}{\milli\second}$ in AEBS and $\SI{100}{\milli\second}$ in SDEC), and thus, the methods can be used for real-time out-of-distribution detection.
 

\begin{table}[!ht]
	\centering
	\caption{Execution times.}
	\label{tab:real-time}
	\begin{tabular}{c c c c c c c}
	\hline
	\multicolumn{1}{m{0.05\columnwidth}}{} 
	& \multicolumn{1}{m{0.05\columnwidth}}{\centering $N$} 
	& \multicolumn{1}{m{0.08\columnwidth}}{\centering $\min$ (ms)}
	& \multicolumn{1}{m{0.08\columnwidth}}{\centering $Q_1$ (ms)}
	& \multicolumn{1}{m{0.08\columnwidth}}{\centering $Q_2$ (ms)}
	& \multicolumn{1}{m{0.08\columnwidth}}{\centering $Q_3$ (ms)}
	& \multicolumn{1}{m{0.08\columnwidth}}{\centering $\max$ (ms)} \\
	\hline
	AEBS &N/A & 3.48 & 3.85 & 3.91 & 3.96 & 4.20 \\
	\hline
	SDEC &N/A & 2.20 & 2.37 & 2.45 & 2.56 &  3.31\\
	\hline
	\multirow{3}{*}{VAE} &5 & 18.64 & 18.68 & 18.71 & 18.73 & 18.81 \\
	& 10 & 37.26 & 37.38 & 37.41 & 37.45 & 38.11 \\
	& 20 & 75.12 & 75.19 & 75.22 & 75.68 & 78.83 \\
	\hline
	\multirow{3}{*}{SVDD} &5 & 2.18 & 2.19 & 2.19 & 2.2 & 2.26 \\
	& 10 & 2.17 & 2.19 & 2.19 & 2.2 & 2.29 \\
	& 20 & 2.17 & 2.22 & 2.23 & 2.23 & 2.37 \\
	\hline
\end{tabular}
\end{table}
	\section{Related Work}
\label{sec:related}
Verification and assurance of CPS with machine learning components is considered in~\cite{seshia2016towards} in a broader context of verified artificial intelligence. The challenges discussed in~\cite{seshia2016towards} include the integration of design-time and runtime methods to address the undecidability of verification in complex systems and environment modeling. Out-of-distribution detection can be used with recovery and reconfiguration techniques to complement design-time verification. Focusing on design-time techniques, an approach to identify regions of the input space that lack training data and potentially larger errors is presented in~\cite{gu2019towards}. The approach could be adapted to predict at runtime if new inputs are from regions covered during training or not. Compositional falsification of CPS with machine learning components is introduced in~\cite{dreossi2017compositional} and demonstrated with a simulated AEBS. The approach is applied at design-time for identifying executions that falsify temporal logic specifications and also identifies regions of uncertainty where additional analysis and runtime monitoring is required.
A related approach for simulation-based adversarial test generation for autonomous vehicles with machine learning components is presented in~\cite{tuncali2018simulation}. The technique is also used at design-time to increase the reliability of autonomous CPS and can provide additional training data for out-of-distribution detection.

Detection of out-of-distribution examples in neural networks has received considerable attention especially in the context of classification tasks in computer vision~\cite{hendrycks2016baseline}. Correctly classified examples tend to have greater maximum softmax probabilities than erroneously classified and out-of-distribution examples. An approach for improving detection by training anomaly detectors is proposed in~\cite{hendrycks2019oe}. An approach for reducing the number of false alarms of out-of-distribution image detection by adding small perturbations to the input is presented in~\cite{Liang2017EnhancingTR}. The idea is similar to randomly sampling from the latent space of the VAE that can also improve the reliability of the detection. Such detection techniques do not take into consideration the dynamical behavior of CPS and can exhibit large number of false alarms. 

An approach which aims to detect novelties based on the reconstruction error of an autoencoder for a single input is presented in~\cite{Richter2017SafeVN}. The approach is used for safe visual and LIDAR-based navigation of mobile robots. A similar approach using VAEs is proposed in~\cite{mcallister2019robustness} to estimate the uncertainty for a collision prediction task for a robot car. As discussed in~\cite{Richter2017SafeVN, mcallister2019robustness}, out-of-distribution detection and, in general, uncertainty estimation is an important \replaced{research}{researh} direction for providing a more robust detection. In our approach, we leverage inductive conformal prediction and anomaly detection for developing a detection algorithm that has well-calibrated false alarm rate~\cite{laxhammar2015inductive} and we compute the nonconformity scores in real-time using VAE and SVDD neural networks.

Conformal anomaly detection is used for sequential anomaly detection of time trajectories in~\cite{laxhammar2013online}. A nonconformity measure based on nearest neighbors and Hausdorff distance provides good performance and robustness but it cannot be used for high-dimensional trajectories at runtime. Inductive techniques have been used for change-point detection in~\cite{volkhonskiy2017inductive,smith2014anomaly} but do not consider high-dimensional inputs and real-time detection. A deep $k$-NN method based on conformal prediction that uses a nonconformity measure computed using the representation learned by each layer of the DNN classifier is presented in~\cite{papernot2018deep}. 
The method computes confidence metrics for evaluating the robustness of the DNN classifier and out-of-distribution detection. 
Estimating the confidence and the uncertainty of neural network prediction is an important problem. In addition to conformal prediction, the problem is addressed using various techniques for computing a confidence or trust score~\cite{devries2018learning,jiang2018trust}. A promising research direction is to combine these with conformal prediction for improving the robustness and reliability of neural networks. 

	\section{Conclusions}
In this work, we demonstrated a method for out-of-distribution detection in learning-enabled CPS. The method is based on inductive conformal prediction and anomaly detection but uses VAEs and SVDD to learn models to efficiently compute the nonconformity of new inputs relative to the training set and enable real-time detection of high-dimensional out-of-distribution inputs. Our evaluation is based on an AEBS and an SDEC implemented in an open source simulator for self-driving cars. The results demonstrate very small number of false positives and detection delay while the execution time is comparable to the execution time of the original LECs. Detection of out-of-distribution inputs does not necessarily imply large errors. An important question is \replaced{whether}{weather} it is possible to identify inputs that result in large errors. Evaluation with real-world data sets is also part of our future work.

	\nocite{*}
	\bibliographystyle{IEEEtran}

	
\end{document}